\documentclass[10pt,twocolumn,letterpaper]{article}

\usepackage{cvpr}
\usepackage{times}
\usepackage{epsfig}
\usepackage{graphicx}
\usepackage{amsmath}
\usepackage{amssymb}
\usepackage{caption}
\usepackage{subcaption}
\usepackage{multirow}
\usepackage{float}

\usepackage[breaklinks=true,bookmarks=false]{hyperref}

\cvprfinalcopy 

\def\cvprPaperID{****} 

\ifcvprfinal\fi
\begin{document}

\title{Crop Lodging Prediction from UAV-Acquired Images of Wheat and Canola using a DCNN Augmented with Handcrafted Texture Features}

\author{Sara Mardanisamani$^{\star1}$, Farhad Maleki$^{\star1}$, Sara Hosseinzadeh Kassani$^1$, Sajith Rajapaksa$^1$, Hema Duddu$^2$\\ Menglu Wang$^2$, Steve Shirtliffe$^2$, Seungbum Ryu$^2$, Anique Josuttes$^2$, Ti Zhang$^2$, Sally Vail$^3$\\ Curtis Pozniak $^2$, Isobel Parkin$^3$, Ian Stavness$^1$ and Mark Eramian$^1$\\\and 
\strut\\
$^1$Department of Computer Science\\
University of Saskatchewan\\
Saskatoon, SK, Canada\\
\texttt{\small sara.samani@usask.ca, farhad.maleki@usask.ca}\\
\texttt{\small mark.eramian@usask.ca}
\and
\strut\\
$^2$Department of Plant Sciences\\
University of Saskatchewan\\
Saskatoon, SK, Canada\\
\and 
$^3$Agriculture and Agri-food Canada\\
Saskatoon Research and Development Center\\
Saskatoon, SK, Canada\\
}

\maketitle
\thispagestyle{empty}

\begin{abstract}
Lodging, the permanent bending over of food crops,  leads to poor plant growth and development. Consequently, lodging results in reduced crop quality, lowers crop yield, and makes harvesting difficult. Plant breeders routinely evaluate several thousand breeding lines, and therefore, automatic lodging detection and prediction is of great value aid in selection. In this paper,  we propose a deep convolutional neural network (DCNN) architecture for lodging classification using five spectral channel orthomosaic images from canola and wheat breeding trials.  Also, using transfer learning, we trained 10 lodging detection models using well-established deep convolutional neural network architectures. 
Our proposed model outperforms the state-of-the-art lodging detection methods in the literature that use only handcrafted features. In comparison to 10 DCNN lodging detection models, our proposed model achieves comparable results while having a substantially lower number of parameters. This makes the proposed model suitable for applications such as real-time classification using inexpensive hardware for high-throughput phenotyping pipelines. The GitHub repository
at \url{https://github.com/FarhadMaleki/LodgedNet} contains code and models.
	\let\thefootnote\relax\footnote{$^\star$ Co-first authors.}
	\addtocounter{footnote}{-1}\let\thefootnote\svthefootnote
	
\end{abstract}
%
\section{Introduction}
Lodging occurs when plant stems break or bend over so that plants are permanently displaced from their optimal upright position. It is a common problem for many crops, including wheat and canola, and can be caused by external forces, including wind, rain, or hail~\cite{wu2016new}, and morphological factors, such as thin or weak stem structures.\par 
Multiple studies on rice, wheat, and oats have shown that lodging can cause grain yield loss and deterioration in seed quality \cite{ma2012comparisons}.  In addition, lodging can cause problems for harvest operations often resulting in  increasing the demand for grain drying, which raises production costs. In most crops, severe lodging results in as much as a 50\% yield reduction \cite{berry2004understanding}. For plant breeders, it is important to identify lodging-resistant varieties from thousands of experimental plots. Therefore, automatic lodging detection methods from overhead images are valuable. In addition, crop insurance claims following wind/hail storms currently require manual assessment of crop damage. Lodging detection from aerial drone-acquired images could provide a faster and more accurate assessment of the area and severity of lodging within a field.\par
Researchers have developed some field methods to cope with lodging and determine the best way to obtain the maximum harvestable product. However, manual in-the-field assessment requires estimating the part of the field that is lodged and the degree of lodging plants relative to their vertical axis. These manual measurements are time-consuming and costly~\cite{berry2002lodging}, as well as quite subjective. To monitor crop lodging within an entire field, unmanned aerial vehicles (UAVs) may automatically collect high-resolution aerial images to detect the lodging in a simple, flexible, cost-effective way~\cite{yang2017spatial}. Then image analysis techniques can be used for automatic lodging detection.\par
We propose a deep convolutional neural network architecture that couples handcrafted and learned features to detect lodging. Previously in the literature, only methods based on handcrafted features have been used for lodging detection from images~\cite{Rajapaksa2018classification,wang2018unsupervised,yang2015wheat,yang2017spatial}. Deep convolutional neural networks (DCNNs) have been successfully applied to a wide range of image classification tasks~\cite{he2016deep,huang2017densely,iandola2016squeezenet,krizhevsky2012imagenet,simonyan2014very,szegedy2016rethinking}. However, to the best of our knowledge, they have not been used for lodging detection. In general, methods that only rely on handcrafted features often achieve lower accuracy in comparison to deep CNN models. They are also sensitive to noise. DCNN-based models, on the other hand, often disregard research on problem-specific handcrafted features. Furthermore, they need a substantial amount of training data to achieve high accuracy. \par
To avoid the shortcomings of both approaches and to benefit from their strengths, in this paper, we propose LodgedNet, an architecture that uses a DCNN-based model together with two texture feature descriptors: local binary patterns (LBP) and gray-level co-occurrence matrix (GLCM) for crop lodging classification.
%
%
%
We also developed 10 DCNN-based models using well-established architectures. LodgedNet is designed to offer rapid training and prediction time while achieving accuracy comparable to that of the 10 DCNN-based models.
To the best of our knowledge, there is no DCNN-based model in the literature applied to lodging detection. This work offers a comprehensive study of CNN architectures (including LodgedNet) for lodging detection.\par 
%
The rest of the paper is organized as follows. Section~\ref{sec:RelatedWorks} presents the related work on lodging detection. In Section~\ref{sec:MaterialsandMethods}, we describe the lodging datasets for wheat and canola and present the proposed architecture. Section~\ref{sec:ExperimentalResults} presents experimental results. Section~\ref{sec:Discussion} discusses the results and the utility of the proposed architecture for similar applications in agriculture. Finally, Section~\ref{sec:Conclusion} ends the paper with a short summary and conclusion.\par
\section{Related Works}\label{sec:RelatedWorks}
Rajapaksa et al.~\cite{Rajapaksa2018classification} used handcrafted features and a support vector machine (SVM) for lodging classification using data obtained from drone imagery.  They extracted features using a gray level co-occurrence matrix (GLCM), local binary patterns (LBP), and Gabor filters. Then they used an SVM to classify the feature vectors extracted from images of lodged and non-lodged plots. Their method was designed for grayscale images and they used information only from a single image channel to predict lodging.\par
Wang et al.\cite{wang2018unsupervised} proposed a method for lodging detection using pixel information obtained from wheat plot images taken by drones. They calculated nine colour features based on pixel values. In addition, they obtained 13 features from the nine colour features using the ENvironment for Visualizing Images (ENVI) software. 
They then used a thresholding approach to discriminate lodged pixels. Their approach solely relies on thresholding and pixel values and disregards spatial information and high correlation among pixel neighbourhoods. This makes the result sensitive to noise.\par
Yang et al.~\cite{yang2017spatial} proposed a spectral and spatial hybrid image classification method to detect rice lodging using images taken by drones. They obtained spatial information, including height data, using the IBM 3D construction algorithm and texture features. In addition to spatial information of the field, they extracted spectral information of each pixel using single feature probability (SFP). Then the extracted features were used by a  decision tree classifier and a maximum likelihood classifier to detect lodging. Using RGB images and texture features, they achieved an accuracy of 88.14\%. Incorporating pixel-wise spatial information, they achieved the accuracy values of 90.76\% and 96.17\% using the maximum likelihood classifier and the decision tree classifier, respectively. However, extracting spatial features requires using extra equipment that makes this approach expensive and time-consuming. \par
Yang et al.~\cite{yang2015wheat} (a different group of researchers) used satellite data (RADARSAT-2) to detect lodging in wheat fields. They extracted a set of sensitive polarimetric features and backscattering intensity features from five consecutive RADARSAT-2 images throughout the entire growing season to detect lodging. Using this approach they were able to identify lodged fields. However, plant breeders require lodging detection on the much smaller scale of their breeding plots which typically rectangular plots measuring a few meters on each side. Satellite imagery cannot provide the required resolution for assessment of individual breeder plots of this size.

In this research, we design an architecture based on handcrafted and DCNN features. Combining these types of features have been proposed previously~\cite{hosseini2018feeding,nguyen2018combining,wang2014cascaded}. Using an image along with Gabor filters extracted from that image as input to the network, Hosseini et al.~\cite{hosseini2018feeding} achieved a higher accuracy compared to several traditional and CNN-based models.
Wang et al.~\cite{wang2014cascaded}, using a cascaded approach based on combining a CNN model and handcrafted features, proposed a computationally efficient model for counting the number of cells undergoing mitosis.
Nguyen et al.~\cite{nguyen2018combining} used a CNN model and multi-level local binary pattern (MLBP) for presentation attack detection (PAD) in face recognition.
They combined features extracted from a CNN-based model and the multi-level local binary pattern (MLBP) method to build a support vector machine classifier. Their model achieved a higher accuracy compared to previous PAD methods.

To the best of our knowledge, all of the published work on crop lodging detection have been developed using handcrafted features tailored to one or a few specific types of crops. Although models based on handcrafted features are often computationally efficient and applicable even in situations where we do not have access to a large number of training examples, these models often have been designed for a specific crop type and might not achieve a comparable accuracy when applied to other crop types. Furthermore, adjusting a handcrafted feature to a different task often is not straightforward and requires further research.
Deep convolutional neural networks (DCNN), on the other hand, have proven to be an effective approach in machine vision. However, DCNN-based models often require a large amount of training data. In this paper, we propose a model that benefits from the strengths of both handcrafted features and also DCNNs, and avoids the shortcomings of these approaches.\par
%
\section{ Materials and Methods}\label{sec:MaterialsandMethods}
\subsection{Data Set Description}
The wheat and canola datasets used in this study were obtained from two breeding field trials. Plot images were taken with \ifcvprfinal{a Draganfly X4P quad-copter (Draganfly Innovations Inc., Saskatoon, SK, Canada) carrying a MicaSense RedEdge camera (Micasense Inc. Seattle, WA, USA)  in the summer of 2016.}\else{[details of imaging equipment redacted for anonymous review].}\fi\ 
This camera captures images with five spectral channels: red, blue, green, near infrared, and red-edge.
Agisoft Photoscan  (Agisoft LLC, St. Petersburg, Russia) was used to stitch images and obtain a high-resolution orthomosaic image of each field. The ground resolution of the obtained images is approximately 15 to 26 mm/pixel for each band.  Images were taken at a height of $20$ meters for canola and $30$  meters for wheat.
Figure 1 shows the orthomosaic image extracted from the red, green, and blue channels.\par 
\begin{figure}
	\centering
	\includegraphics[width=\linewidth]{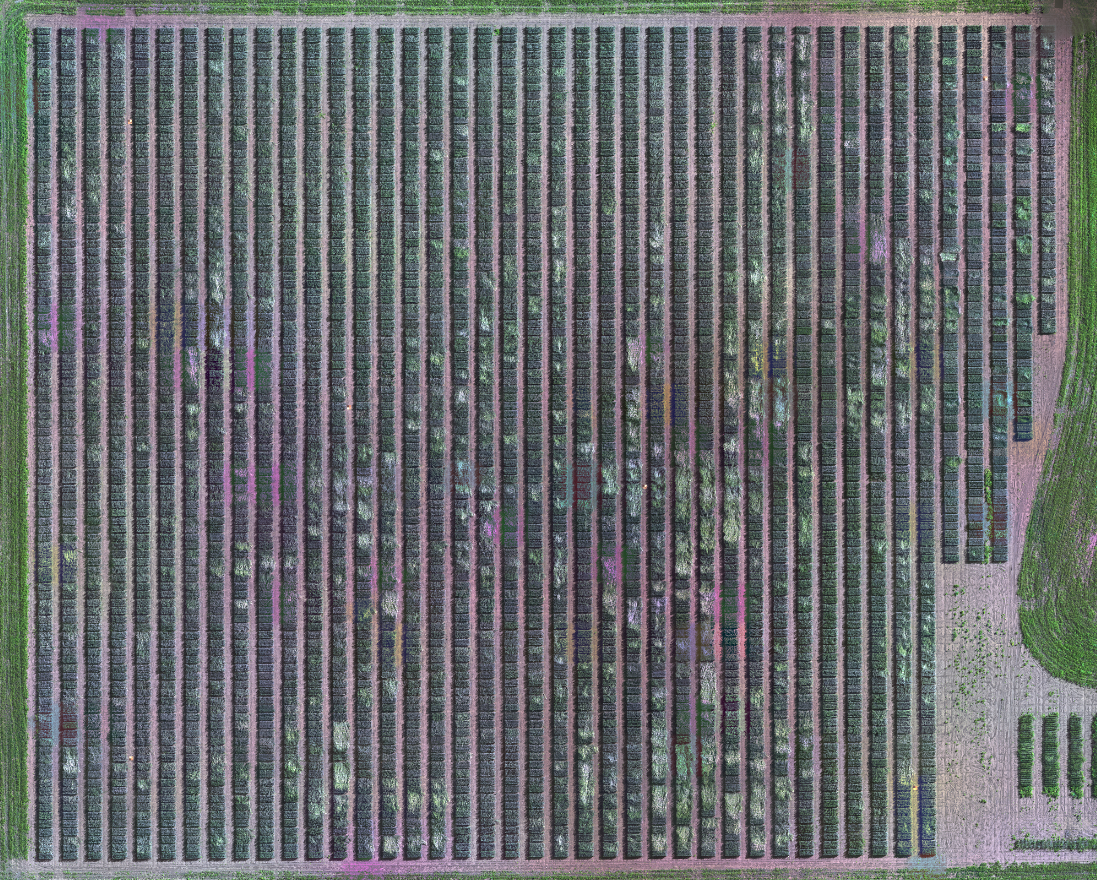}	
	\caption{An RGB orthomosaic image for a wheat  trial.}
	\label{Figure1}
\end{figure}
%
\subsection{Data}
\label{sec:TrainingandTestingSamples}
Two orthomosaic images for wheat ($9492\times8340$ pixels) and seven orthomosaic images for canola ($9492\times8340$ pixels) are used in this study. Each field is organized into several columns, and each column is divided into several small plots (see Figure~\ref{Figure1}). Table~\ref{TAB:LodgedNonLodgedNum} represents the number of extracted plots for wheat and canola categorized as lodged and non-lodged.
\par 
\begin{table}[!tbhp]
	\centering
	\begin{tabular}{|l|c|c|c|}
		\hline
		Samples       & Non-lodged & Lodged & Total \\\hline\hline
		Wheat  & 285        & 180    & 465   \\\hline
		Canola & 1170       & 468    & 1638 \\\hline
	\end{tabular}
	\caption{Number of samples for wheat and canola datasets for each class.}
	\label{TAB:LodgedNonLodgedNum}
\end{table}
We extracted image samples  with dimensions  $60\times100$ and $118\times348$ pixels from wheat and canola plots, respectively (the different sample sizes are due to the wheat and canola plots being of different physical size).  Ground truth labels of ``lodged'' or ''not lodged' were provided by a crop agronomist and a plant scientist. Figure 2 illustrates lodged and non-lodged samples from wheat and canola plots.\par
We randomly split each dataset into training and test sets. We used about 80\% of the images for training and the remaining 20\% for the test set. Training data were split randomly with 80\% of training samples used for a training set and the remaining 20\% for a validation set. 
Table~\ref{TAB:TrainValidTestNum} shows the number of samples in the training, validation, and test sets.\par
\begin{table}[h]
	
	\centering
	\begin{tabular}{|c|c|c|c|c|}
		\hline
		
		\multirow{2}{*}{Data} & \multicolumn{2}{c|}{{Wheat}}                                            & \multicolumn{2}{c|}{{Canola}} \\
		\cline{2-5}
		& \multicolumn{1}{c|}{{NL}} & {L} & \multicolumn{1}{c|}{{NL}} & {L} \\\hline\hline
		{Train}        & 187                                      & 113                                  & 754                                      & 300                                  \\ \hline
		{Validation}   & 48                                       & 28                                   & 188                                      & 75                                   \\ \hline
		{Test}         & 50                                      & 39                                   & 228                                      & 93                                   \\ \hline
	\end{tabular}
	\caption{The number of non-lodged (NL) and lodged (L) samples in the training, validation, and test sets.}
	\label{TAB:TrainValidTestNum}
\end{table}
\begin{figure}[h]
	\centering
	\begin{subfigure}{0.44\linewidth}
		\centering
		\includegraphics[width=0.85\linewidth]{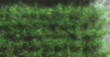}
		\caption{}
	\end{subfigure}%
	\begin{subfigure}{0.44\linewidth}
		\centering
		\includegraphics[width=0.85\linewidth]{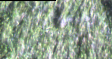}
		\caption{}
	\end{subfigure}
	\begin{subfigure}{\linewidth}
		\centering
		\includegraphics[width=0.8\linewidth]{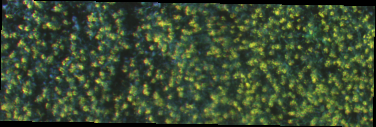}
		\caption{}
	\end{subfigure}
	\begin{subfigure}{\linewidth}
		\centering
		\includegraphics[width=0.8\linewidth]{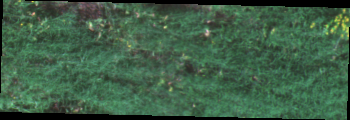}
		\caption{}
	\end{subfigure}%
	\caption{RGB images for wheat and canola:  a) a non-lodged wheat plot; b) a lodged wheat plot; c) a non-lodged canola plot; and d) a lodged canola plot.}
	\label{Figure2}
\end{figure}

\begin{figure*}[h]
	\centering
	\includegraphics[scale=0.5]{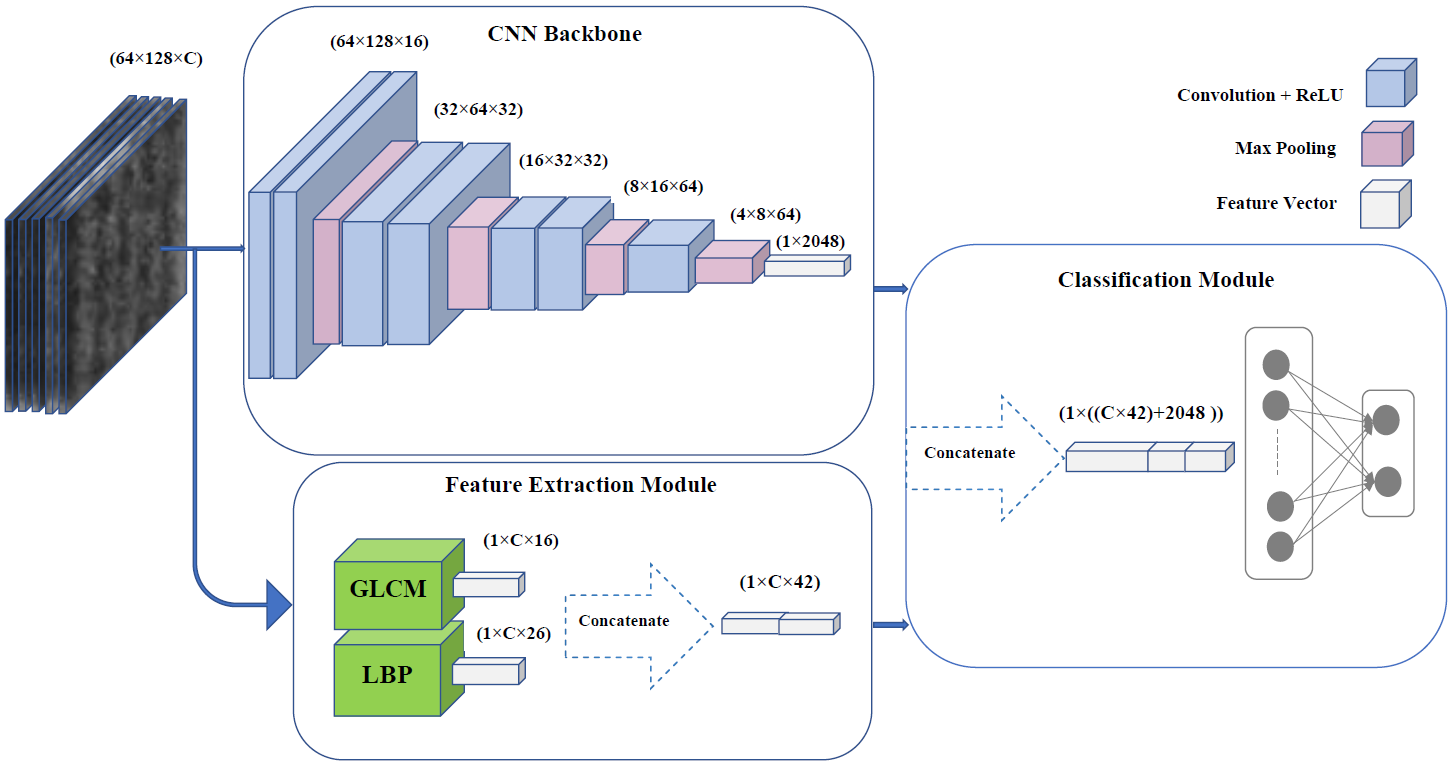}
	\caption{A schematic diagram of LodgedNet, the proposed architecture. All convolutional and fully connected layers are followed by a dropout layer with a ratio of 0.5 except convolutional layers one and three; $C$ represents the number of image channels.}
	\label{FIG:LodgedNetArchitecture}
	
\end{figure*}
%
\subsection{The proposed architecture}
\label{SEC:Architecture}

Convolutional neural networks are a well-established approach in computer vision contributing to the success of applications in image segmentation, object detection, and image classification~\cite{mohanty2016using, vgg, inception}.
In this section, we propose LodgedNet, which is a DCNN-based architecture for lodging detection. LodgedNet, as illustrated in Figure~\ref{FIG:LodgedNetArchitecture}, consists of three main components: a CNN backbone, a texture feature extraction module, and a classification module.  The CNN backbone consists of seven convolutional layers, with 16, 16, 32, 32, 32, 32, and 64 filters respectively. We used filters of size three, paddings of size one, and Rectified Linear Unit (ReLU) activation functions.  A \textit{SpatialDropout}~\cite{tompson2015efficient} layer with a dropout rate of 0.5 was used after each convolutional layer, except for the first and third layers, to prevent overfitting.   Max-pooling layers are used before the third, fifth, and seventh convolutional layers, and after the seventh convolutional layer (See Figure \ref{FIG:LodgedNetArchitecture}).
The texture extraction module extracts LBP and GLCM features from each channel of the input images.  These features are image properties related to second-order statistics that account for the spatial inter-dependencies of pixels at specific relative positions.
The extracted texture features and the features generated from the last layer of the CNN module are flattened and concatenated to be used as input to the classification module.
The classification module consists of two fully connected layers, respectively of 128 and two neurons, followed by a final softmax layer. To further prevent overfitting one dropout layer~\cite{srivastava2014dropout} was applied before, and another after, the first fully connected layer. A dropout rate of 0.5 was used for these layers. This architecture has been designed to achieve high accuracy while having a small number of parameters to facilitate training and deployment on low-cost hardware.\par
\section{Experimental Results}\label{sec:ExperimentalResults}
\subsection{Feature Extraction Module}
In this section, we describe the model specification for LodgedNet as well as the implementation details.
%
%
The LBP ~\cite{ojala2002multiresolution} and GLCM ~\cite{haralick1973textural} texture extractor methods used in the proposed architecture were applied with the following parameters. They are the same as those selected by Rajapaksa et al. \cite{Rajapaksa2018classification} in their work which uses only GLCM and LBP for lodging classification. We extracted 16 \textit{contrast} texture features with GLCM using the normalized and asymmetric 2D co-occurrence matrices for each channel of an input image using four orientations (0$^\circ$, 90$^\circ$, 135$^\circ$, 180$^\circ$) and four distances (1px, 2px, 4px, 5px).
For LBP, we used 8-bit rotationally-invariant uniform local binary patterns and  constructed 10-bin LBP histograms to be used as features.  In addition, we extracted 16-bin LBP variance histograms and concatenated them with the 10 LBP features as recommended by Ojala et al. ~\cite{ojala2002multiresolution}. For each image channel, 16 GLCM features and 26 LBP features were extracted. Then these features were concatenated to be used as an input to the classification module. The LBP and GLCM features obtained using the ``greycomatrix'', ``greycoprops'', and ``local\_binary\_pattern'' functions from ``scikit-image'' library.
\subsection{Data Augmentation and Training}
It is common practice to perform data augmentation to further increase the generalizability of trained models. We employed a number of transformations for data augmentation. We resized the original images while preserving their aspect ratios and then center-cropped the resized images to achieve fixed-size inputs of $64\times 128$ for LodgedNet and fixed-size inputs of $224\times 224$ for all well-established models except Inception-V3, where a fixed-size input of $299\times 299$ is required. A cropped image undergoes a vertical and a horizontal flipping each independently with a probability of 0.5. Each image was normalized by subtracting the mean and dividing by the standard deviation, where mean and standard deviation were calculated as the per-channel mean and standard deviation of the training and validation data.\par
%
%
All DCNN architectures were implemented in Python using the PyTorch package Version 1.0 on a Intel Core i7-5930K 3.5 GHz processor and NVIDIA GTX 1080 Ti with 11 GB graphical processing unit (GPU) and 32 GB RAM.\par
LodgedNet was trained on both wheat and canola samples from scratch, while for the 10 well-established models, a  version of the models pretrained with ImageNet data~\cite{krizhevsky2012imagenet} were used. To make a fair comparison, for the 10 well-established models we redefined the classifier component of each model to match that of LodgedNet's classification module.
The pretrained models were further trained using two approaches. In the first approach, we let all parameters of the models be further trained using the lodging datasets. In the second approach, we froze all parameters other than the parameters of the classifier component, which were learned using the lodging datasets. 
In all experiments, an Adam optimizer with a learning rate of 0.001, and $\beta_1=0.9$, and $\beta_2=0.999$ were used. Also, we used a batch size of 16 and number of epochs equal to 50.\par 
Table \ref{TAB:PretrainedOurs} shows the results of LodgedNet and the 10 well-established models for wheat and canola datasets, respectively. Also, Table \ref{TAB:HandOurs} illustrates the results of two state-of-the-art methods using handcrafted features.  These results represent the first applications of DCNNs to the problem of lodging detection.\par

\begin{table*}[tbhp]
	\centering
	\begin{tabular}{|l|c|c|l|c|c|c|c|}
		\hline
		\multicolumn{1}{|c|}{\textbf{Architecture}}            & \textbf{Wheat}               & \textbf{\begin{tabular}[c]{@{}c@{}}Wheat\\ (FW)\end{tabular}} & \multicolumn{1}{c|}{\textbf{Canola}} & \textbf{\begin{tabular}[c]{@{}c@{}}Canola\\ (FW)\end{tabular}} & \textbf{\begin{tabular}[c]{@{}l@{}}Number of\\  Parameters\end{tabular}} & \textbf{\begin{tabular}[c]{@{}c@{}}Prediction Time\\Canola (ms) \\ $\boldsymbol{\mu\pm\sigma}$\end{tabular}}& \textbf{\begin{tabular}[c]{@{}c@{}}Prediction Time \\ Wheat (ms) \\ $\boldsymbol{\mu\pm\sigma}$\end{tabular}}\\ \hline
		VGG19~\cite{simonyan2014very}                                 & 98.68\%                      & 97.75\%                                                                    & 98.75\%                              & 99.06\% &    143,667,240 & 5.75 $\pm$ 0.92  &   9.99 $\pm$ 0.00                                                                  \\ \hline
		VGG16~\cite{simonyan2014very}                                 & 98.68\%                      & 96.62\%                                                                    & 97.19\%                              & 98.44\% &  138,357,544 &4.99 $\pm$ 0.75 & 4.99 $\pm$ 0.00                                                                \\ \hline
		AlexNet~\cite{krizhevsky2012imagenet}                               & \multicolumn{1}{l|}{98.85\%}   & 97.75\%                                                                    & 89.09\%                              & 90.65/\% & 61,100,840  &   2.95 $\pm$ 0.65 & 2.83 $\pm$ 0.37                                                                   \\ \hline
		ResNet101~\cite{he2016deep}                              & \multicolumn{1}{l|}{100\%} & 95.50\%                                                                    & 98.44\%                              & 98.44\% & 44,549,160 & 34.94 $\pm$ 4.67 & 41.66 $\pm$ 6.87                                                                      \\ \hline
		Inception-V3~\cite{szegedy2016rethinking}                          & 100\%                        & 97.75\%                                                                          & 99.68\%                              & 100\%  & 27,161,264 &  36.66 $\pm$ 6.42 & 33.13 $\pm$ 4.52                                                                    \\ \hline
		ResNet50~\cite{he2016deep}                               & 97.70\%                      & 100\%                                                                      & 99.06\%                              & 99.06\%  & 25,557,032 &  18.88 $\pm$ 3.77 & 21.66 $\pm$ 3.73                                                                    \\ \hline
		DensNet201~\cite{huang2017densely}                            & 98.85\%                      & 100\%                                                                      & 98.75\%                              & 99.06\% & 20,013,928  & 81.39 $\pm$ 7.21 & 88.33 $\pm$ 14.62                                                                    \\ \hline
		DensNet169~\cite{huang2017densely}                            & 100\%                        & 100\%                                                                      & 98.75\%                              & 99.37\%  &14,149,480 &   67.12 $\pm$ 7.08 &  71.66 $\pm$ 10.67                                                                   \\ \hline
		ResNet18~\cite{he2016deep}                               & 100\%                        & 100\%                                                                      & 99.37\%                              & 98.44\%   & 11,689,512  &   7.18 $\pm$ 1.05 &    7.49 $\pm$ 0.5                                                               \\ \hline
		SqueezeNet~\cite{iandola2016squeezenet}                            & 97.70\%                      & 98.87\%                                                                    & 98.75\%                              & 98.75\%  & 1,235,496 &   10.46 $\pm$ 1.73 & 9.99 $\pm$ 5.77                                                                  \\ \hline
		\textbf{LodgedNet} & \textbf{97.70\%}                    & -                                                                          & \textbf{99.06\%}                            & -  &   \textbf{332,306}  & \textbf{2.99} $\boldsymbol{\pm}$ \textbf{0.31} &  \textbf{3.49} $\boldsymbol{\pm}$ \textbf{0.49}                                                                 \\ \hline
	\end{tabular}
	\caption{Comparison between the proposed model and the 10 lodging detection models developed based on well-established architectures. Red, green, and blue channels were used as input images for wheat and canola; FW denotes the case of training with frozen weights; $\mu$ and $\sigma$ denote average and standard deviation of the prediction time in milliseconds.  The models are presented in descending order of number of parameters.}
	\label{TAB:PretrainedOurs}
\end{table*}

\begin{table*}[tbhp]
	\centering
	\begin{tabular}{|l|c|c|c|c|l|l|}
		\hline
		\multicolumn{1}{|c|}{\textbf{Case}}    & \multicolumn{3}{c|}{\textbf{Wheat}}                     & \multicolumn{3}{c|}{\textbf{Canola}}                                                \\ \hline
		\multicolumn{1}{|c|}{\textbf{Methods}} & \textbf{RGB} & \textbf{Five Channel} & \textbf{Rededge} & \multicolumn{1}{l|}{\textbf{RGB}} & \textbf{Five Channel}  & \textbf{Rededge}       \\ \hline
		GLCM-based~\cite{Rajapaksa2018classification}                                    & -            & -                     & 84.94\%          & -                                 & \multicolumn{1}{c|}{-} & 72.86\%                \\ \hline
		LBP-based~\cite{Rajapaksa2018classification}                                     & -            & -                     & 96.77\%          & -                                 & \multicolumn{1}{c|}{-} & 90.54\%                \\ \hline
		\textbf{LodgedNet}                        & \textbf{97.70\%}      & \textbf{97.33\%}                 & -                & \multicolumn{1}{l|}{\textbf{99.06\%}}      & \textbf{99.38\%}                & \multicolumn{1}{c|}{-} \\ \hline
	\end{tabular}
	\caption{Comparison between our method and Rajapaksa's method \cite{Rajapaksa2018classification} (state-of-the-art in the literature for lodging classification) which used two handcrafted features for wheat and canola datasets on only one image channel.}
	\label{TAB:HandOurs}
\end{table*}


\section{Discussion}\label{sec:Discussion}

As can be observed from Table~\ref{TAB:PretrainedOurs} and \ref{TAB:HandOurs}, LodgedNet achieved comparable results to the 10 DCNN-based classifiers and outperformed Rajapaksa et al.'s \cite{Rajapaksa2018classification} state-of-the-art model for both crops.\par
Table~\ref{TAB:PretrainedOurs} shows the number of parameters for the architectures used in this study. LodgedNet has a substantially lower number of parameters compared to all of the other DCNN-based classifiers. For example, the number of parameters in VGG19  is about 432 times more than that of LodgedNet. Among the DCCN-based models, SqueezeNet has the smallest number of parameters, which is almost 3.7 times more than that of LodgedNet. Despite the substantially smaller number of parameters,  LodgedNet was among the top three methods with the highest test accuracy for the canola dataset and only misclassified two test samples from the wheat dataset.\par
Furthermore, the small number of parameters makes deploying LodgedNet on low-cost hardware possible, leading to near real-time inference. Using drone imagery along with mobile or low-cost portable computers, models such as LodgedNet that are computationally less demanding will be accessible to a wide range of agricultural applications.
%
%
However, a sound comparison of the inference time for  LodgedNet and the 10 DCNN-based models is difficult to achieve due to several factors:  LodgedNet works with both three- and five-channel images, the input image dimensions of the tested networks differ, and the DCNN backbones of the 10 lodging detection models are highly-optimized implementations from the \textit{torchvision} package of PyTorch. As a coarse comparison of these models, we used three-channel (RGB) images to estimate the inference time as the average of inference times for test samples. As depicted in Table~\ref{TAB:PretrainedOurs}, LodgedNet comes second to AlexNet in prediction time while achieving comparable accuracy for wheat and a substantially higher accuracy for canola (about 10\% higher) when compared to AlexNet.\par
In this paper, we focused on proposing an architecture that is not computationally demanding and makes using handcrafted features possible. We used handcrafted texture feature extractors for lodging detection. However, the use of handcrafted features is not limited to lodging detection and such features are available for various application domains. Using these features can help to increase model accuracy, more specifically in domains where there is not a large amount of training data available. \par

In this study, we used five-channel images taken by drones for lodging detection. 
 The results of our experiments showed that even in the absence of red edge and near-infrared channels, lodging detection can be performed with high accuracy. However, red-edge and near-infrared channels might contribute to achieving higher accuracy in other agricultural applications. We suggest using these channel data, which are supported by LodgedNet, for other agricultural applications.\par
\subsection{Future Work and Limitations}
LodgedNet used two handcrafted feature extraction approaches, namely GLCM and LBP. However, LodgedNet is not limited to using these two feature extractors. Extending LodgedNet to use other handcrafted features is suggested as future research.  Because lodging tends to make the texture of the overhead plot views more directional, we expect that using Gabor filters~\cite{jain2000filterbank} alongside GLCM and LBP might lead to higher accuracy.\par
In this paper, we used LodgedNet for lodging classification. However, the proposed architecture can be used to tackle other image classification problems where there are handcrafted features available from previous research. Although models that only rely on handcrafted features often lead to lower accuracy and higher sensitivity to noise, in comparison to their DCNN-based counterparts, incorporating these handcrafted features in an architecture similar to LodgedNet could potentially help with improving accuracy.  This approach could be of more value where there are relatively few training samples available.\par
A limitation of our assessment of prediction time is that it did not take into account the time to compute the LBP and GLCM features.  This took about 30ms on average for a plot image input to LodgedNet because they were computed with a single-threaded CPU algorithm but this is not directly comparable to the DCNN forward pass prediction times since those computations were performed on a GPU.  A parallelized implementation would greatly reduce the LBP and GLCM feature computation time.  Regardless of this cost, the lower forward pass computation time of LodgedNet should reduce training time because the LBP and GLCM features can be pre-computed and do not need to be extracted repeatedly for each training epoch or during the tuning of hyper-parameters.\par
One limitation of the transfer learning used for the 10 DCNN-based models is that they have been trained using RGB images and therefore the transfer learning using images that use extra channels such as red edge and near infrared cannot be used if we choose to use transfer learning. Considering the number of parameters for these models, training them from scratch requires a substantially large number of training samples; otherwise, considering their large capacity, they tend to overfit and the trained models are less likely to be generalizable to unseen samples. LodgedNet, however, can be used with a variable number of input image channels because the number of channels, $C$, is a hyper-parameter.  Our extensive use of \textit{SpatialDropout}~\cite{tompson2015efficient} and regular dropout layers~\cite{srivastava2014dropout} provides resilience to overfitting.\par
%
%
\section{Conclusion}\label{sec:Conclusion}
In this paper, we used DCNNs to address crop lodging classification. We trained 10 models based on well-established DCNN architectures pre-trained on ImageNet data. We then proposed a new architecture, LodgedNet, that utilizes both DCNN and handcrafted feature extractors to build a lodging prediction network.
Our comprehensive study of lodging prediction compared 11 DCNN architectures, including LodgedNet. LodgedNet, as well as all of the other DCNN architectures tested, outperformed current state-of-the-art lodging detection models using only handcrafted features.  LodgedNet's prediction accuracy compares favourably with the 10 other architectures tested while having about one quarter of the number of trainable parameters compared to SqueezeNet, which is the next smallest network. 
The fewer number of parameters in LodgedNet accelerates training and inference time. It also facilitates deploying LodgedNet on low cost hardware.\par
LodgedNet can be used for lodging detection in high-throughput plant phenotyping scenarios. Such pipelines will be critical in the future search for higher yield crops varieties needed to feed a growing population.


\section{Acknowledgements}

This research was enabled thanks to funding from the Canada First Research Excellence Fund.

{\small
	\bibliographystyle{ieee}
	\bibliography{25}
}

\end{document}